\documentclass[runningheads]{llncs}
\usepackage[T1]{fontenc}
\usepackage{graphicx}
\usepackage{subfigure}
\usepackage{stfloats}
\usepackage{floatrow}
\usepackage{amsmath}

\newcommand{\upcite}[1]{\textsuperscript{\textsuperscript{\cite{#1}}}}
\begin{document}
%
%\title{Contribution Title\thanks{Supported by organization x.}}
\title{Efficiently Trained Low-Resource Mongolian Text-to-Speech System Based On FullConv-TTS}
%\author{ZiQi Liang\inst{1}\orcidID{0000-1111-2222-3333} \and
%Second Author\inst{2,3}\orcidID{1111-2222-3333-4444} \and
%Third Author\inst{3}\orcidID{2222--3333-4444-5555}}

\author{Ziqi Liang}

\authorrunning{Ziqi Liang et al.}
\institute{University of Science and Technology of China, China \\
\email{tymlzq@mail.ustc.edu.cn}\\}
\maketitle              % typeset the header of the contribution
\begin{abstract}

Recurrent neural networks (RNNs) have become a standard modeling technique for sequential data and are used in novel text-to-speech models. However, training a TTS model which includes RNN components requires powerful GPU performance and takes a long time. In contrast, CNN-based sequence synthesis techniques can significantly reduce the training time of a text-to-speech model while guaranteeing a certain performance due to its high parallelism.
We propose a novel text-to-speech system based on deep convolutional neural networks that does not employ any RNN components and is a two-stage training end-to-end TTS model. Meanwhile, we improve the robustness of our model by a series of data enhancement methods, such as time warping, frequency masking and time masking, for the low resource problem.
We propose a novel text-to-speech system based on deep convolutional neural networks, which does not employ any RNN components (recurrent units) and is a two-stage training end-to-end TTS model. Also, to address the low resource problem of lacking labeled data, we improve the robustness of our model by a series of data enhancement methods such as time warping, frequency masking and time masking.
The final experimental results show that a TTS model using only CNN components can reduce the training time while ensuring the quality and naturalness of the synthesized speech compared to using mainstream TTS models, such as Tacotron2 and the vocoder Hifigan. Our solution achieved 12th place in the NCMMSC2022-MTTSC\upcite{ref_article0}.

\end{abstract}
\section{INTRODUCTION}

At present, the speech synthesis technology of mainstream languages such as Chinese and English has developed relatively maturely, and the speech synthesis of low-resource languages has gradually attracted more and more attention.

Mongolian is the most famous and widely spoken language among the Mongolian people. Worldwide, there are about 6 million users. At the same time, Mongolian is also the main national language of China's Inner Mongolia Autonomous Region\upcite{ref_article1}. Therefore, the study of Mongolian-oriented speech synthesis technology is of great significance to the fields of education, transportation, and communication in minority areas.

Traditional speech synthesis methods mainly include speech synthesis techniques based on waveform splicing and statistical parametric acoustic modeling (e.g. Hidden Markov Model). With the development of deep learning, DNN-based acoustic models and vocoders are gradually being widely used. 

For acoustic modeling, the more popular one is end-to-end acoustic modeling, which mainly adopts Encoder-Decoder structure to learn the mapping of text and acoustic parameter pairs directly, among which the more representative ones are Tacotron1/2\upcite{ref_article7}, and the transformer-based variant models\upcite{ref_article8}.

The Tacotron\upcite{ref_article7} series of models are decoded with the output of the previous moment as the input of the next moment for acoustic parameter prediction. This autoregressive decoding structure greatly limits the real-time performance of speech synthesis and does not fully utilize the GPU parallel computing resources.
In order to improve the decoding speed, researchers further propose speech synthesis models based on non-autoregressive acoustic modeling such as FastSpeech\upcite{ref_article9} and FastSpeech2\upcite{ref_article10}. The non-autoregressive acoustic models can take a given text as input and output the whole sequence of acoustic parameters in parallel, without relying on the acoustic parameters obtained from the decoding of historical moments.

For vocoder research, a neural network-based vocoder has also been proposed to directly model the speech sample points. The neural vocoder directly learns the mapping relationship between speech parameters and speech waveform sample points, which significantly improves the fidelity of the synthesized speech. There are also vocoders based on autoregressive structures for speech waveform sample point prediction, such as Wavenet, and non-autoregressive structures with high fidelity and fast generation speed, such as Hifi-GAN\upcite{ref_article11}, MelGAN\upcite{ref_article12}, etc.

However, using a model such as Tacotron\upcite{ref_article7} as an acoustic model has a drawback that it uses many recursive components with high training cost, which is demanding on GPU computing resources.

In this paper, we propose the Mongolian Text-to-Speech model "FullConv-TTS", which is a sequence2sequence model based entirely on CNN modules. Experiments show that the synthesized audio of the system proposed in this paper, while ensuring a certain sound quality and fidelity, significantly reduces the training time; secondly, Guided Attention Loss likewise enables the attention module to be trained quickly.

In addition, in order to cope with the current situation of low-resource Mongolian language data, we introduce data enhancement means such as time mask and frequency mask to obtain new audio samples and increase the training samples.

\subsection{Related Work}

\subsubsection {Neural TTS}
In recent years, deep neural network-based Text-to-speech models, such as those using RNN structures\upcite{ref_article5,ref_article6}, and the more common DeepVoice1/2, WaveNet, etc., have achieved high quality results.

Later, people gradually began to study to reduce the reliance on handcrafted features, models rely only on mel or linear spectrograms, do not use other speech parameters such as fundamental frequency, resonance peak parameters, etc., and focus only on the spectral representation of the audio signal, of which the more typical is Tacotron.

WaveNet, as a back-end vocoder, is the first model that introduces DNN into the back-end classifier, which also does not use any RNN unit component, but a full convolution-based vocoder. Its input is the speech spectrogram output from the acoustic model, and its output is the speech waveform, a spec2wav structure. Our FullConv-TTS is essentially a front-end model, a Text2spec structure, and finally the waveform is synthesized by the vocoder.

\section{Methodology}
The waveform and spectrogram of a piece of audio are usually interconverted by the short-term Fourier transform (STFT) and the linear mapping of the inverse STFT. In speech synthesis, we usually consider its amplitude spectrum, combined with Griffifin\&Lim algorithm (GLA) to synthesize relevant audio. In addition, mel spectrograms obtained by using a set of mel filter banks are also common in speech synthesis tasks. We use the mel spectrum in the data preprocessing stage, and use a series of data enhancement methods to obtain new enhanced samples. In the training stage, the mel spectrum array of the random mask is used as the training set and mixed with the original mel spectrum in a certain proportion, and sent to The model is trained hybrid. In the time dimension, we select a time frame time frame every four time frames for dimensionality reduction processing, and the mel spectrum is normalized and sent to the Text2Mel model for training.

The model structure for our experiments is as follows:

\begin{figure}[ht]
    \centering
    \includegraphics[width=12cm]{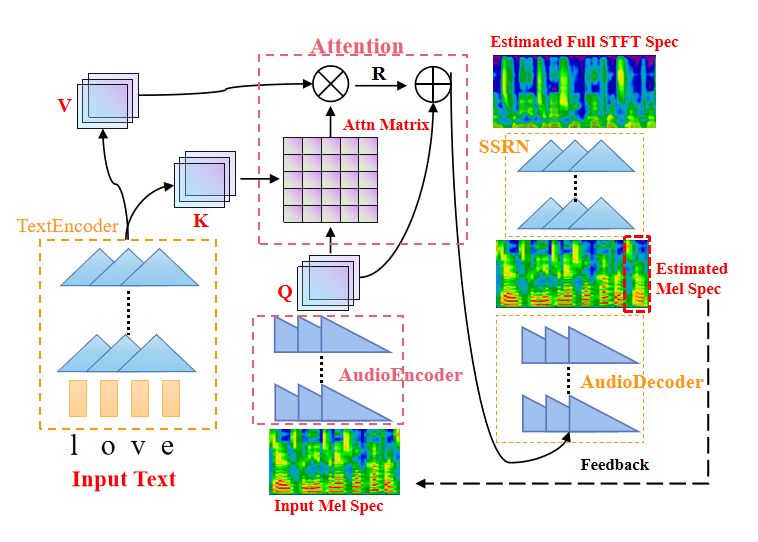}
    \caption{FullConv-TTS Network architecture} 
    \label{model_artiture}
\end{figure}

We use a TTS system based entirely on CNN, and one of the advantages of abandoning the use of recursive unit components is that compared with the TTS system based on RNN components, the model training speed is faster while ensuring a certain degree of fidelity and naturalness, and the requirements for GPU are also lower. Not high, it is friendly to individual contestants; secondly, we use a two-stage synthesis strategy. Compared with the operation of directly synthesizing mel spectrogram or STFT spectrogram from text, we first synthesize a low-resolution coarse mel spectrogram from text, and then synthesize it. For the high-resolution complete STFT spectrogram, the waveform file is finally obtained through a vocoder. 

Therefore, we use the following two networks as acoustic models to synthesize the spectrum. 1) Text2Mel: synthesis of mel spectrogram from input text; 2) spectrogram super-resolution network (SSRN): synthesis of complete STFT spectrogram from coarse mel spectrogram.

\subsection{Text2Mel Network}

We use this module to synthesize a coarse mel spectrogram from the text. The network is composed of four sub modules: Text Encoder, Audio Encoder, Audio Decoder and Attention. Text Encoder first encodes the input text sentence 
$L=[ l_{1}, l_{2},..., l_{N}]$ into K and V, with shape $d \times N$. where d is the dimension of the encoded character. On the other hand, Audio Encoder will convert the coarse Mel spectrum of the voice with the length of T $(dim: F\times T)$ encoded as matrix Q $(dim: d \times T)$.

% \begin{gather}

%   $ K, V = TextEncoder(L) $\\
%   $ Q = AudioEncoder(S_{1:F,1:T}) $ \\
%   f_{FGM-E} = C(A(Cat(C(f_{s}^{i}), C^{2}(f_{e}^{i-1})))
% \end{gather}

\begin{equation}
K, V = TextEncoder(L) 
\end{equation}
\begin{equation}
Q = AudioEncoder(S_{1:F,1:T}) 
\end{equation}
\begin{equation}
A = Softmax(K^{T}Q / \sqrt{d})
\end{equation}

Attention matrix $A$ is used to evaluate the correlation between the $n$-th character $l_{n}$ and the $t$-th  mel spectrogram $S_{1:F,t}$. At the same time, the attention module will pay attention to the character $l_{n+1}$ in the following time and encoded to the $n$-th line of $V$. Therefore, the matrix R as the subsequent Mel spectrum is defined as:
\begin{equation}
R = attn(Q,K,V) := V A
\end{equation}

Then, the encoded audio matrix $Q$ and matrix $R$ are spliced into $R'$, which is used as the input of audio decoder to estimate Mel spectrum.
\begin{equation}
\hat{Y}_{1:F,2:T+1} = AudioDecoder(R'), R'=Concat[R,Q]
\end{equation}

The error of prediction results and ground truth is evaluated by loss function $\mathcal{L}_{hiera}({Y}_{1:F,2:T+1}|{s}_{1:F,2:T+1})$, which consists of $\mathcal{L}_{1}$ $Loss$ and binary divergence function $\mathcal{L}_{spec}$:
\begin{equation}
    \begin{aligned}
        \mathcal{L}_{spec}(\mathbf{Y}|\mathcal{S}) &= \mathbf{E}_{ft}[-\mathcal{S}_{ft}log(\frac{\mathbf{Y}_{ft}}{{S}_{ft}}) - (1-{S}_{ft})log(\frac{1-{Y}_{ft}}{1-{S}_{ft}})] \\
        &=\mathbf{E}_{ft}[-\mathcal{S}_{ft}\hat{Y}_{ft} + log(1+e^{\hat{Y}_{ft}})]         
    \end{aligned}
\end{equation}

\begin{equation}
\mathcal{L}_{hiera} = \mathcal{L}_{spec}(\mathbf{Y}|\mathcal{S}) + \mathbf{E}[| \mathbf{Y}_{ft} - \mathcal{S}_{ft}|]
\end{equation}

\subsubsection{Details} TextEncoder consists of a character embedding and several 1-D no-causal convolution layers. AudioEncoder and AudioDecoder are composed of 1-D causal convolution layers. These convolutions should be causal, because the output of AudioDecoder is fed back to the input of AudioEncpder in the synthesis phase.

\subsection{Spectrogram Super-resolution Network}

In the second stage of synthesis, we use SSRN to further synthesize the complete spectrum from the coarse Mel spectrum. For frequency up sampling, we can increase the number of channels in the 1-D convolution network, and for upsampling on the time axis, we can increase the sequence length from $T$ to four times the original length by twice deconvolution.

Since we do not consider online data processing, all convolutions in SSRN are non causal. The loss function used by this module is the same as that of Text2mel in Phase I.

\subsection{Data Enhancement}
\subsubsection{SpecArgument}
Facing the current situation of low resources of Mongolian language data, we consider using data augmentation to supplement the training samples and improve the model robustness, and the common methods are adding natural noise or artificial noise, volume enhancement, speed enhancement, etc. The advantage of noise augmentation is that it can make the model more robust and applicable to more scenarios, but the disadvantage is that it requires a large amount of noisy data, and insufficient data will affect the generalization ability.

In this paper, we use SpecAugment, which simply performs the three operations of Time Warping, Frequency Masking, and Time Masking on the speech spectrogram.

\begin{figure}[htbp]
\centering
\subfigure[Mel Spectrogram.]{
\begin{minipage}[t]{0.5\linewidth}
\centering
\includegraphics[width=6cm]{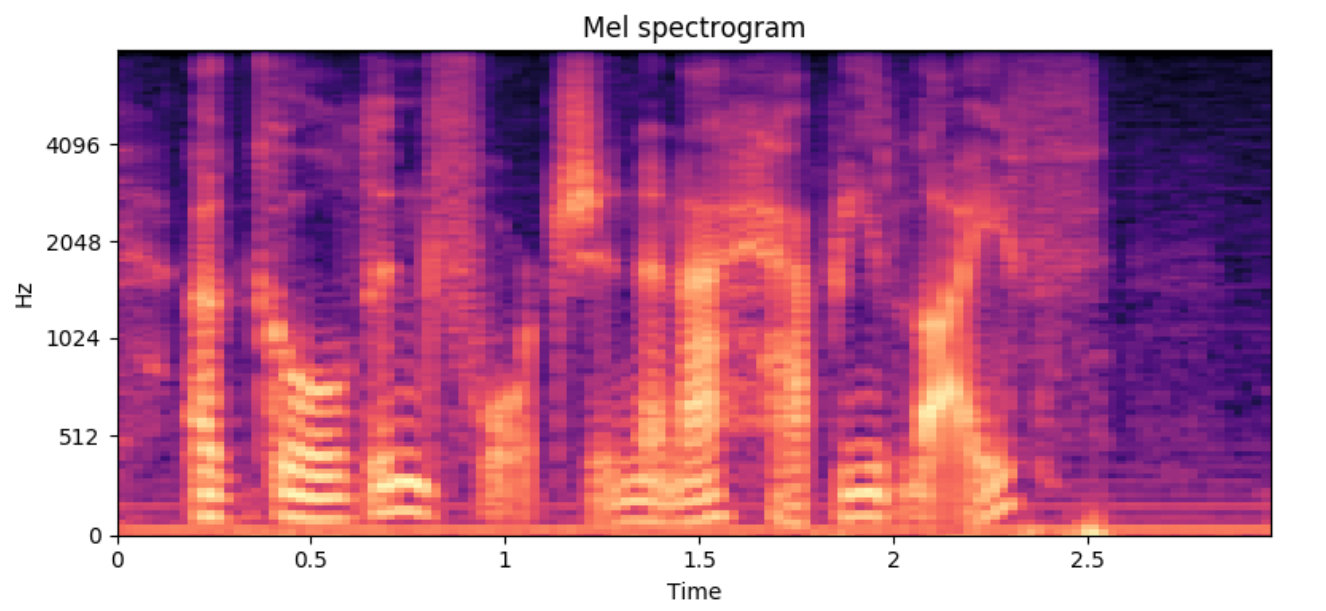}
\end{minipage}%
}%
\subfigure[Mel Spectrogram Masked.]{
\begin{minipage}[t]{0.5\linewidth}
\centering
\includegraphics[width=6cm]{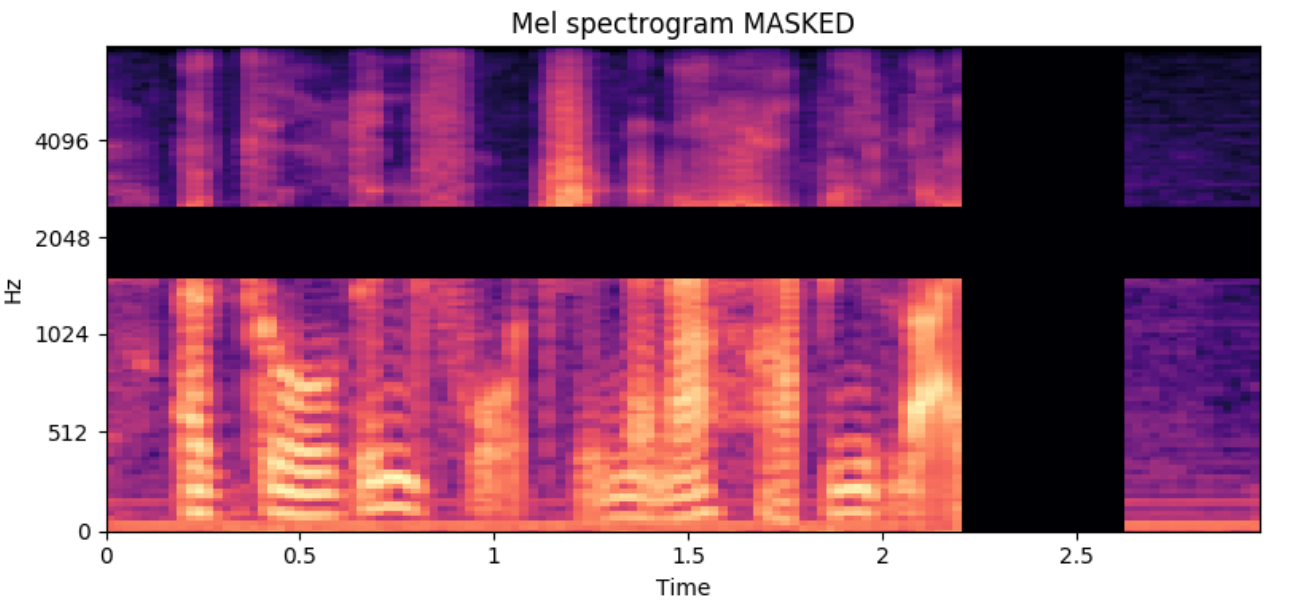}
\end{minipage}%
}%
\centering
\caption{Mel Spectrogram by SpecArgument}
\end{figure}

\begin{itemize}
\item[$\bullet$] \textbf{Time Warping}: A data point is randomly selected on the time axis of the speech spectrogram and randomly distorted to the left or right at a certain distance, with the distance parameter randomly chosen from a uniform distribution of time distortion parameters from point 0 to along that line.
\end{itemize}

\begin{itemize}
\item[$\bullet$] \textbf{Frequency Masking}: The part of the frequency channel domain [f0, f0+f) is masked in its entirety, where f is chosen from a uniform distribution of the parameters from point 0 to the frequency mask, f0 is chosen from (0, v-f), and v is the number of channels in the frequency dimension.
\end{itemize}

\begin{itemize}
\item[$\bullet$] \textbf{Time Masking}: The spectrogram data mask of t consecutive time steps [t0, t0+t] interval, where t0 is chosen from [0,T-t] interval, T is the length of audio duration obtained by framing means, and likewise t is taken from 0 to the uniform distribution of time mask parameters. 
\end{itemize}

The experimental results show that the method does improve the training speed, which is due to the fact that no further data conversion is required between the waveform data to the spectrogram data, and it is a direct enhancement of the spectrogram data.

\subsubsection{Spectrogram-Resize}

In addition to SpecArgument, Spectrogram-Resize data enhancement\upcite{ref_article19} is used to supplement data by compressed or stretched source spectrograms to alleviate the low-resource problem of insufficient training corpus. This method is simple and easier to implement, and does not require complex signal processing knowledge.

\begin{figure}
\centering
\subfigure[Resize ratio < 1]{
\begin{minipage}[t]{1\textwidth}
\includegraphics[width=1\textwidth]{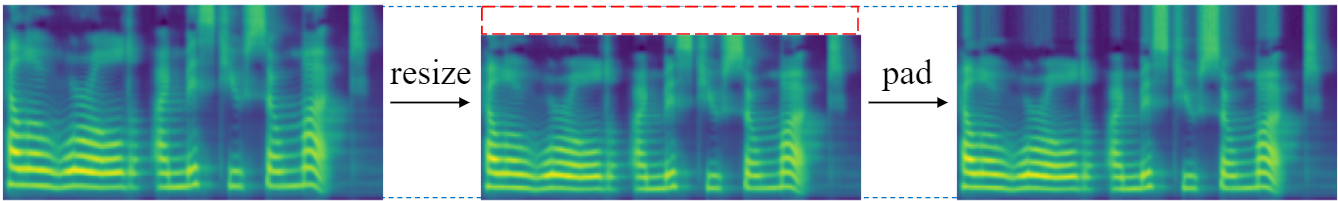} \\
\end{minipage}
}
\centering
\subfigure[Resize ratio > 1]{
\begin{minipage}[t]{1\textwidth}
\includegraphics[width=1\textwidth]{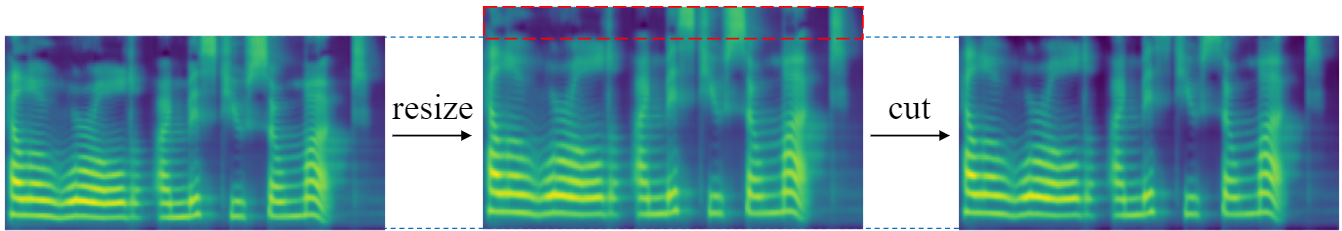}
\end{minipage}
}
\caption{Vertical Spectrogram-Resize}
\end{figure}

Spectrogram-Resize data enhancement operations are as follows: extract Mel spectrogram from audio waveforms, and perform vertical frequency bin axis resize or horizontal time axis resize operation on the spectrogram. Using the frequency bin axis as an example, as shown in Fig.3, the melspectrogram is first adjusted to a specific proportion by using bilinear interpolation, and then the melspectrogram is padded or cut to an original shape. If the ratio is less than 1, it produces audio with a lower pitch and a closer formant distance, and vice versa. We train the enhanced audio, and the model can better learn the content information of the speaker in the audio. In addition to the operations on the vertical frequency bin axis, the resize operation can also be performed on the horizontal time axis.

We use SpecAugment to solve the problem of overfitting and model generalization due to the lack of sufficient training data. However, it should be noted that the introduction of data augmentation turns the overfitting problem into an underfitting problem, and although the validation effect on the training set decreases, the effect on the test set does improve, which proves the merit of SpecAugment in improving the generalization of the model.

\section{Guided Attention}
\subsection{Guided Attention Loss}
Due to the correspondence between the order of text characters and audio clips, the attention module in Text-to-speech needs to pay extra attention to the word alignment between different languages, in addition to the fact that the reading step of text characters is often performed linearly in time by default.

We use guided attention loss, which can make the attention matrix distributed near the diagonal, and set a penalty term if the attention matrix is distributed far from the diagonal, which means that characters are loaded in random order and not linearly.

\begin{equation}
\mathcal{L}_{attn}(A) = \mathbf{E}_{attn}[\mathcal{A}_{attn}\mathcal{W}_{attn}]
\end{equation}

$\mathcal {W}_{attn}$ in the above equation is $1-exp[-(n/N-t/T)]^{2}/2\mathbf{g}^{2}$, where $\mathbf{g}$=0.2, $\mathcal{L}_{attn}(A)$ as an auxiliary loss function, is updated iteratively with the main loss function $\mathcal{L}_{hiera}$. In our experiments, if we add bootstrap attention loss as an auxiliary loss function to the objective, it reduces the number of iterations required for training and indirectly reduces the time consumed to train the text-to-speech model.

\subsection{Enhanced Robustness}

During the synthesis phase, the attention matrix A sometimes fails to focus on the correct characters, typically skipping a few letters and repeating the same word twice or more. To make the system more robust, we heuristically modify the attention matrix to be distributed close to the diagonal by some simple rules. This approach sometimes alleviates such problems.

\section{Experiments and Results}

The NCMMSC2022-MTTSC dataset was recorded by a professional female announcer whose native language is Mongolian. The entire recording process was recorded in a standard recording studio at Inner Mongolia University using Adobe Audition software. The announcer followed our text script and read aloud sentence by sentence. In addition, a volunteer supervises the recording process and asks the announcer to re-record if there are any murmurs or unreasonable pauses during the recording process.

\subsection{Experimental Setup}

We train on the officially provided training set and use the validation set to evaluate the model performance and select the model with the best results for inference testing. The experimental device is a 1080Ti with 12G memory. we train Text2mel and SSRN for about 15h and 30h respectively.

\begin{table}[]\centering
\caption{Parameters Set.}\label{tab1}
\begin{tabular}{c|c}
\hline
\textbf{Parameter}    & \textbf{Value}                   \\ \hline
STFT win function     & Hanning Win                      \\ \hline
STFT win length       & 1024                             \\ \hline
STFT win shift        & 256                              \\ \hline
Sampling rate         & 22050 Hz                         \\ \hline
Adam $\alpha$, $\beta_{1}$, $\beta_{2}$, $\xi$   & 2e-4, 0.5, 0.9, 10e-6       \\ \hline
Dimension e, d and c  & 128, 256, 512                    \\ \hline
STFT spectrogram size & 80 × T (T depends on audio clip) \\ \hline
Mel spectrogram size  & 80 × T (T depends on audio clip) \\ \hline
\end{tabular}
\end{table}

For simplicity, we train Text2Mel and SSRN independently and asynchronously using different GPUs. all network parameters are initialized using a Gaussian initializer. Both networks are trained by the ADAM optimizer. When training the SSRN, we randomly extract short sequences of length T = 64 for each iteration to save memory usage. To reduce disk access, we reduce the frequency of creating parameters and saving the model to only once per 2K iterations.

\subsection{Result}
The system evaluation is intended to address both naturalness and intelligibility of the synthesized speech. The evaluation metrics include Naturalness Mean Opinion Score (N-MOS), Intelligibility Mean Opinion Score (I-MOS), and Word Error Rate (WER). three items.

The test results of our submitted inference audio are as given in Fig.4
\begin{figure}[ht]
    \centering
    \includegraphics[width=10cm]{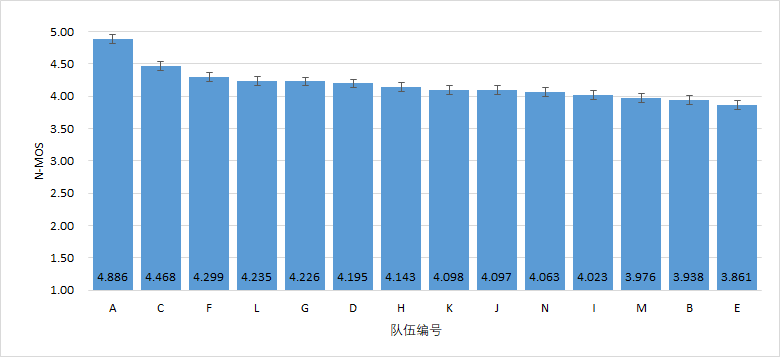}
    \includegraphics[width=10cm]{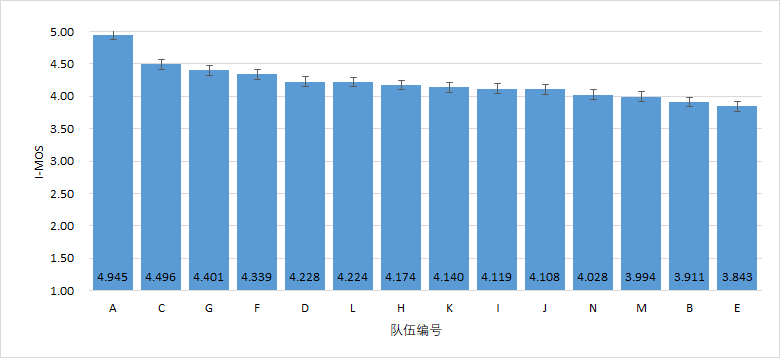}
    \caption{N-MOS And I-MOS Scores} 
    \label{N-MOS}
\end{figure}

From Fig.4, we can see that our FullConv-TTS solution ranks 12th among all participating teams in terms of N-MOS and I-MOS metrics (TeamNumber B), where A is the MOS score of bonafide audio. It can be seen that our proposed system has N-MOS score of 3.938 and I-MOS score of 3.911, which has the great advantage of significantly reducing the training time and is not very demanding on the experimental equipment, while ensuring an acceptable degree of naturalness and intelligibility.

\begin{table}[]\centering
\caption{Comparison of MOS.}\label{tab3}
\begin{tabular}{c|c}
\hline
\textbf{Method}              & \textbf{MOS}                    \\ \hline
Tacotron+Griffin-Lim (iteration 877k*)   & $3.07 \pm 0.12$                  \\ \hline
Tacotron2+HifiGAN                        & $4.01 \pm 0.08$                  \\ \hline
Fastspeech2+HifiGAN                      & $3.93 \pm 0.09$                  \\ \hline
FullConv-TTS                             & $3.92 \pm 0.08$                  \\ \hline
\end{tabular}
\end{table}

\begin{table}[]\centering
\caption{Comparison of Training Times.}\label{tab3}
\begin{tabular}{c|c}
\hline
\textbf{Method}                      & \textbf{Training Times}       \\ \hline
Tacotron + Griffin-Lim (iteration 877k*)   & ~288 hours                   \\ \hline
Tacotron2 + HifiGAN                        & ~168 hours                 \\ \hline
Fastspeech2 + HifiGAN                      & ~75 hours                   \\ \hline
FullConv-TTS                              & ~45 hours                  \\ \hline
\end{tabular}
\end{table}

We conducted several groups of experiments on Mongolian datasets provided by the organizer. Tacotron(Griffin-Lim vocoder), Tacotron2, and Fastspeech2 were used as acoustic models, and HiFiGAN v1 generators were used as vocoders for training on Mongolian datasets. Finally, the trained model is used for joint reasoning, and the experimental results of several groups of schemes are evaluated. The experimental results are shown in Table 2 and 3. Although the audio quality and fidelity of our model are still different from Tacotron2 and the current mainstream end-to-end VITS model, our model is constructed entirely by CNN modules because no RNN component is used. The training and inference speed of the model completely overwhelms other solutions. In addition, Fullcon-TTS inference results are basically the same as fastspeech2 in terms of sound quality and fidelity, but the inference speed is improved by 40\%.

\begin{table}[]\centering
\caption{Comparison of DataEnhancement.}\label{tab3}
\begin{tabular}{c|c|c}
\hline
\textbf{Method}              & \textbf{MOS(w)}     & \textbf{MOS}               \\ \hline
FullConv-TTS + SA (w/o)      & $\approx3.90$    & $\approx3.916$      \\ \hline
FullConv-TTS + SR (w/o)      & /                 & $\approx3.920$           \\ \hline
FullConv-TTS + SA/SR(w/o)    & /                 & $\approx3.923$    \\ \hline
\end{tabular}
\end{table}

We performed ablation experiments on the two data augmentation techniques employed in our solution. The experimental results indicate that training on the augmented dataset leads to an improvement in the audio quality and fidelity of the speech generated by Fullconv-TTS inference, regardless of whether a single data augmentation technique is used or a combination of both amplification and other techniques are used in sequence. Furthermore, the prosodic information is well preserved in the synthesized speech.

\subsubsection{Analysis}
A novel TTS technique based on deep convolutional neural networks and a technique for fast training of Guided attention modules is used in this paper. Although it is certainly less accurate in training compared to the mainstream scheme of using Tacotron2/FastSpeech2 to extract mel spectrograms from text and then training HiFiGan or MelGAN as a vocoder, but limited by the experimental equipment, we could only choose this full convolution-based framework to train the text-to speech model, and the experimental results show that we have significantly reduced the training time consumption while ensuring a certain quality of the synthesized audio.

\section{Conclusion}
The experimental results show that although the audio quality is far from perfect, it can be improved by thoroughly tuning some hyperparameters as well as some techniques. In addition, for the low resource problem, we have considered methods such as migration learning using Universal pre-trained models, but are limited by the lack of computational resources, so they are not considered for now. We believe that this simple neural TTS can be extended to many other uses, as only small models can have commercial landing value and be more easily integrated into the company's speech engine products.

\subsubsection{Acknowledgements} Thanks for the official's Mongolian dataset and consulting services


\end{document}